\pdfoutput=1

\documentclass[11pt]{article}

\usepackage[preprint]{acl}
\definecolor{teaserblue}{RGB}{242, 242, 255}
\usepackage{times}
\usepackage{latexsym}
\usepackage{booktabs}
\usepackage[T1]{fontenc}

\usepackage[utf8]{inputenc}

\usepackage{microtype}

\usepackage{inconsolata}

\usepackage{graphicx}

\usepackage{color}

\usepackage[ruled,linesnumbered]{algorithm2e}

\usepackage{xspace}

\usepackage{tabularx}
\usepackage{makecell}
\usepackage{amssymb}
\usepackage{microtype}

\usepackage{inconsolata}

\usepackage{amsmath}
\usepackage{amsfonts}
\usepackage{amssymb}

\usepackage{booktabs}
\usepackage{multirow}

\usepackage{paralist}

\usepackage{mdwlist}

\usepackage[tikz]{bclogo}
\usepackage{color}
\usepackage{xcolor}
\usepackage{lipsum}
\usepackage{tikz}
\usepackage[edges]{forest}
\definecolor{hidden-draw}{RGB}{20,68,106}
\definecolor{hidden-pink}{RGB}{255,245,247}

\title{A Survey on In-context Learning}

\author{
Qingxiu Dong\textsuperscript{\rm1}, 
Lei Li\textsuperscript{\rm1},
Damai Dai\textsuperscript{\rm1}, 
Ce Zheng\textsuperscript{\rm1},
Jingyuan Ma\textsuperscript{\rm1},
Rui Li\textsuperscript{\rm1},
Heming Xia\textsuperscript{\rm2},\\
\textbf{Jingjing Xu\textsuperscript{\rm3}, 
Zhiyong Wu\textsuperscript{\rm4},
Tianyu Liu\textsuperscript{\rm5},
Baobao Chang\textsuperscript{\rm1},
Xu Sun\textsuperscript{\rm1},
Lei Li\textsuperscript{\rm6}
and Zhifang Sui\textsuperscript{\rm1}}  \\
\textsuperscript{\rm 1} Peking University 
\textsuperscript{\rm 2} The Hong Kong Polytechnic University \\
  \textsuperscript{\rm 3} ByteDance \textsuperscript{\rm 4} Shanghai AI Lab \ \ 
  \textsuperscript{\rm 5} Alibaba Group \textsuperscript{\rm 6} Carnegie Mellon University \\
  \texttt{ dqx@stu.pku.edu.cn,}
   \texttt{szf@pku.edu.cn}
}

\begin{document}
\maketitle
\footnotetext[1]{We list the author contributions and roles in Appendix~\ref{app:authers}.}

\begin{abstract}
With the increasing capabilities of large language models (LLMs), in-context learning (ICL) has emerged as a new paradigm for natural language processing (NLP), where LLMs make predictions based on contexts augmented with a few examples. It has been a significant trend to explore ICL to evaluate and extrapolate the ability of LLMs. In this paper, we aim to survey and summarize the progress and challenges of ICL. We first present a formal definition of ICL and clarify its correlation to related studies. Then, we organize and discuss advanced techniques, including training strategies, prompt designing strategies, and related analysis. Additionally, we explore various ICL application scenarios, such as data engineering and knowledge updating. Finally, we address the challenges of ICL and suggest potential directions for further research. We hope that our work can encourage more research on uncovering how ICL works and improving ICL.
\end{abstract}

\section{Introduction}
\label{sec:intro}

With the scaling of model size and data size~\citep{gpt3, chowdhery2022palm, openai:2023gpt4, Hugo:2023llama, Hugo:2023llama2}, large language models (LLMs) demonstrate the in-context learning (ICL) ability, that is, learning from a few examples in the context. 
Many studies have shown that LLMs can perform a series of complex tasks through ICL, such as solving mathematical reasoning problems~\citep{cot}. These strong abilities have been widely verified as emerging abilities for large language models~\citep{wei2022emergent}. 

 



\begin{figure}[t]
    \centering
    \includegraphics[width=0.45\textwidth]{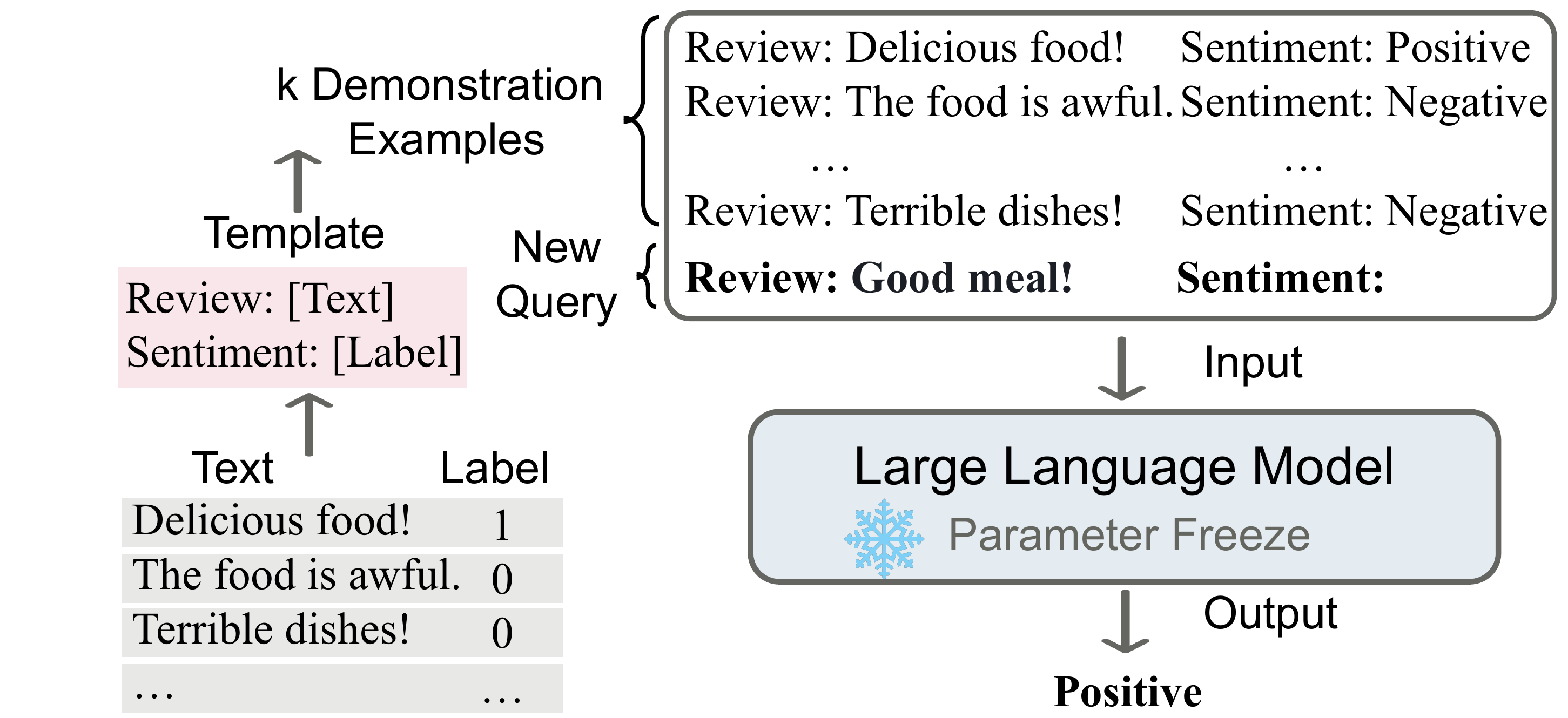}
    \caption{Illustration of in-context learning. ICL requires a prompt context containing a few demonstration examples written in natural language templates. Taking this prompt and a query as the input, large language models are responsible for making predictions.}
    \label{fig:icl}
\end{figure}

The key idea of in-context learning is to learn from analogy. Figure~\ref{fig:icl} gives an example that describes how language models make decisions via ICL.
First, ICL requires a few demonstration examples to form a prompt context. These examples are usually written in natural language templates. 
Then, ICL concatenates a query question and the piece of prompt context together to form the input,  which is then fed into the language model for prediction.
Different from supervised learning, which requires a training stage that uses backward gradients to update model parameters, ICL does not perform parameter updates. The model is expected to learn the pattern hidden in the demonstration and accordingly make the right prediction. 


\tikzstyle{my-box}=[
    rectangle,
    draw=hidden-draw,
    rounded corners,
    text opacity=1,
    minimum height=1.5em,
    minimum width=5em,
    inner sep=2pt,
    align=center,
    fill opacity=.5,
    line width=0.8pt,
]
\tikzstyle{leaf}=[my-box, minimum height=1.5em,
    fill=hidden-pink!80, text=black, align=left,font=\normalsize,
    inner xsep=2pt,
    inner ysep=4pt,
    line width=0.8pt,
]
\begin{figure*}[t!]
    \vspace{-1.0cm}
    \centering
    \resizebox{\textwidth}{!}{
        \begin{forest}
            forked edges,
            for tree={
                grow=east,
                reversed=true,
                anchor=base west,
                parent anchor=east,
                child anchor=west,
                base=left,
                font=\large,
                rectangle,
                draw=hidden-draw,
                rounded corners,
                align=left,
                minimum width=4em,
                edge+={darkgray, line width=1pt},
                s sep=3pt,
                inner xsep=2pt,
                inner ysep=3pt,
                line width=0.8pt,
                ver/.style={rotate=90, child anchor=north, parent anchor=south, anchor=center},
            },
            where level=1{text width=4em,font=\normalsize,}{},
            where level=2{text width=7.5em,font=\normalsize,}{},
            where level=3{text width=6.0em,font=\normalsize,}{},
            where level=4{text width=6.0em,font=\normalsize,}{},
            [
                In-context Learning, ver
                [
                    Training, ver
                    [
                        Pre-training (\S \ref{sec:pretraining})
                        [
                         PICL~\cite{picl}{, }MEND~\cite{Li2023mend}{, }ICLM~\cite{Shi2023iclm}, leaf, text width=45.0em
                        ]
                    ]
                    [
                        Warmup (\S \ref{sec:warmup})
                        [
                            MetaICL~\cite{metaicl}{, }OPT-IML~\cite{optiml}{, }Super-NaturalInstructions~\cite{natural}{, }\\FLAN~\cite{flan}{, }Scaling Instruction~\cite{chung}{, }Self-supervised ICL~\cite{selfsupericl}{, }\\Symbol Tuning~\cite{symboltuning}{, }RICL~\cite{Chu2023Finetune}
                                {, }ICL Markup~\cite{Brunet2023Markup}, leaf, text width=45.0em
                        ]
                    ]
                ]    
                [
                    Inference, ver
                    [
                        Demonstration\\ (\S \ref{sec:demonstration_org})
                        [       
                           Selection \\ (\S \ref{sec:select})
                                [
                                    Unsupervised
                                    [
                                       KATE~\cite{liu2022close}{, }SG-ICL~\cite{kim2022self}{, }Self-Adaptive\\\cite{Wu2022SelfadaptiveIL}{, }PPL~\cite{gonen2022demystifying}{, }MI~\cite{sorensen2022information}{, }\\Informative Score~\cite{li2023supporting}{, }IDS~\cite{qin2023incontext}{,}\\Votek~\cite{su2022selective}
                                        , leaf, text width=29.7em
                                    ]
                                ] 
                                [
                                    Supervised
                                    [
                                      EPR~\cite{rubin2022learning}{, }Q-Learning~\cite{zhang2022active}{, }\\AdaICL~\cite{mavromatis2023examples}{, }Topic~\cite{topic}{, }\\UDR~\cite{udr}
                                        , leaf, text width=29.7em
                                    ]
                                ]
                        ]
                        [
                            Reformatting \\ (\S \ref{sec:reformatting})
                            [   
                            SG-ICL~\citep{kim2022self}{, }Structrured Prompting~\cite{hao2022structured}{, }\\AutoICL~\citep{yang2023autoicl}{, }WICL~\citep{yang2023demonstration}{, }ICV~\citep{liu2024incontext}, leaf, text width=37.3em
                            ]
                        ]
                        [
                            Ordering \\ (\S \ref{sec:order})
                            [ GlobalE\&LocalE~\cite{lu2022order}{,} ICCL~\cite{liu2024lets}
                            , leaf, text width=37.3em
                            ]
                        ]
                    ]
                    [
                        Instruction (\S \ref{sec:instruction})
                        [
                            Instruction Induction~\cite{induct}{, }Self-Instruct~\cite{wang2022self}{, }APE~\cite{zhou2022large}{, }\\Grimoire~\cite{chen2024grimoire}
                                , leaf, text width=45.0em
                        ]
                    ]
                    [
                        Scoring \\ Function (\S \ref{sec:scoring})
                        [
                            Calibrate~\cite{calibrate}{, }Channel Models~\cite{min2022noisy}{, }$k$NN-Prompting~\citep{knnPrompting}, leaf, text width=45.0em
                        ]
                    ]
                ]
                [
                Analysis, ver
                    [
                        Influencing \\Factors  (\S \ref{sec:inf_factors})
                        [ Pre-training \\Stage (\S \ref{sec:inf_factors_pre})
                            [Pre-Training\\ Data
                                [Distribution~\cite{distribution,wies2024learnability}{, }Domain\\\cite{shin2022corpora, Han2023UnderstandingIL}{, }Diversity~\cite{Yadlowsky2023PretrainingDM}, leaf, text width=29.6em
                                ]
                            ]
                            [Model and\\Training
                                [Architecture
                            ~\cite{Ding2023CausalLMIN}{, }Pre-training steps~\cite{wei2022emergent}{, }\\Parameters~\cite{gpt3,wei2022emergent},
                          leaf, text width=29.6em
                                ]
                            ]
                        ]
                        [ Inference\\ Stage (\S \ref{sec:inf_factors_infe})
                            [ Input Labels
                                [
                                Mapping~\cite{ground_truth,Pan2023WhatIL,Tang2023LargeLM}{, }\\Settings~\cite{min2022rethinking},
                          leaf, text width=29.6em
                                ]
                            ]
                            [Demonstration\\ Examples
                                [Diversity and Simplicity~\cite{compositional_generalization}{, }Query Similarity\\\cite{liu2022close, compositional_generalization}{, }Feature bias~\cite{inductive_bias}{, }\\Order~\cite{lu2022order, Zhang2022ActiveES,Liu2023LostIT},
                          leaf, text width=29.6em
                                ]
                            ]
                        ]
                    ]
                    [
                        Learning \\Mechanism (\S \ref{sec:mech}) 
                        [ Functional\\ Modules\\(\S \ref{sec:mech_fun})       
                            [
                            Induction Heads~\cite{olsson2022induction, Bietti2023BirthOA}
                        {, }\\Computational Layers~\cite{label_anchor}{, }Attention Modules~\cite{icl_weight_shifting},
                          leaf, text width=37.3em
                            ]
                        ]
                        [ 
                            Theoretical\\ Interpretation\\(\S\ref{sec:mech_theo})
                            [
                                Bayesian Framework                                      ~\cite{bayesian, topic, jiang2023latent}{, }\\Gradient Descent~\cite{dai2022iclft, Irie2022TheDF, mahankali2023one}{, }\\Others~\cite{garg2022linear, akyurek2022algorithm, trm_as_alg, tr_and_tl}, leaf, text width=37.3em    
                            ]
                        ]
                    ]
                ]  
            ]
        \end{forest}
    }
    \caption{Taxonomy of in-context learning.}
    \label{taxo_of_icl}
\end{figure*}
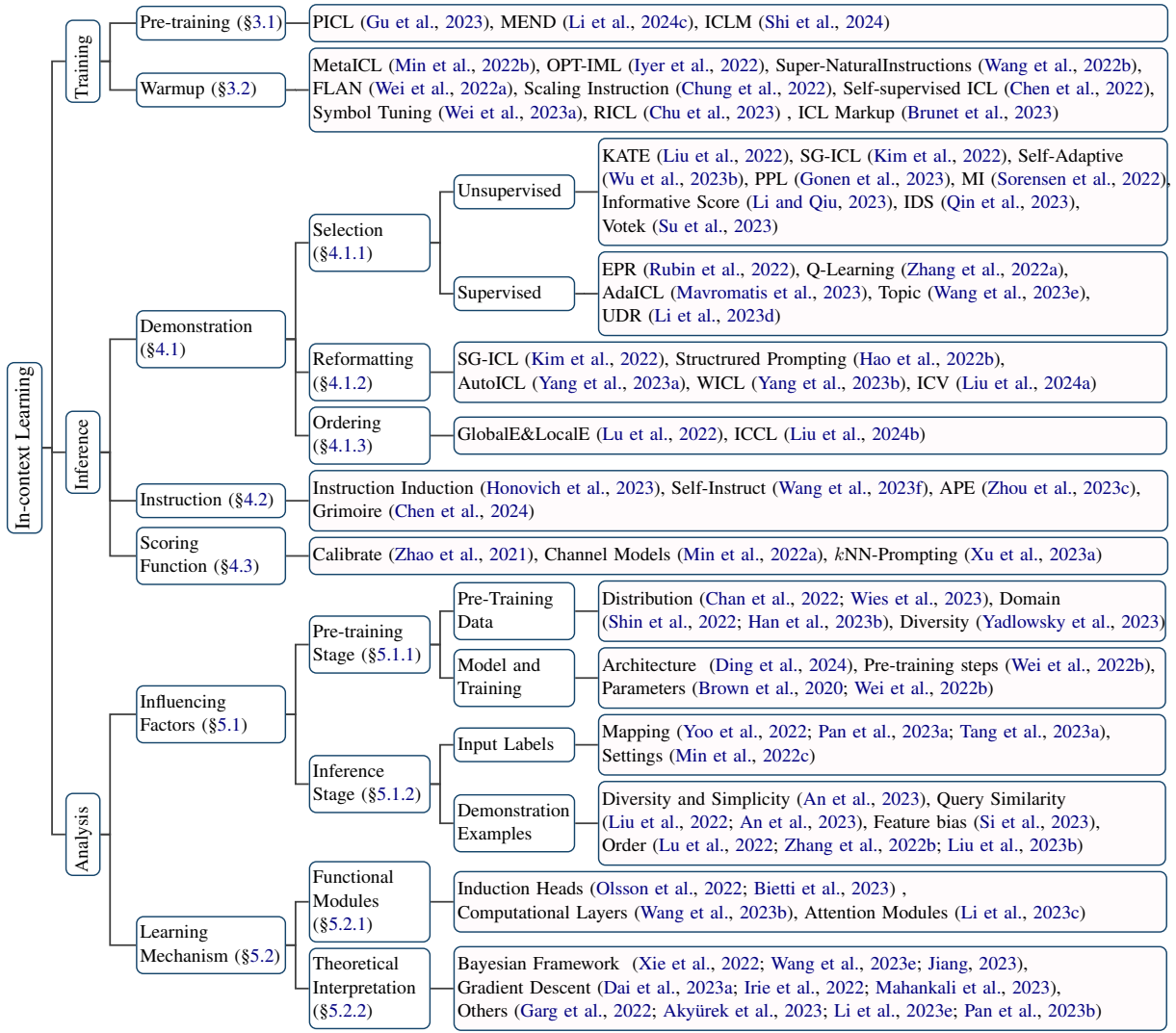

As a new paradigm, ICL has multiple attractive advantages. 
First, since the demonstration is written in natural language, it provides an interpretable interface to communicate with LLMs~\citep{gpt3}.
This paradigm makes it much easier to incorporate human knowledge into LLMs by changing the demonstration and templates~\citep{liu2022close, lu2022order, cot, Wu2022SelfadaptiveIL}. 
Second, in-context learning is similar to the decision process of human beings by learning from analogy~\citep{winston1980learningByAnalogy}. 
Third, compared to supervised training, ICL is a training-free learning framework. 
This could not only greatly reduce the computational costs for adapting the model to new tasks, but also make language-model-as-a-service~\citep{sun2022black} possible and can be easily applied to large-scale real-world tasks.

Despite being promising, there are also interesting questions and intriguing properties that require further investigation in ICL. 
Although a range of vanilla GPT models show excellent ICL capability, several studies have found that this capability can be significantly improved through adaptation during pretraining~\citep{metaicl, Li2023mend}.
Moreover, the performance of ICL is sensitive to specific settings, including the prompt template, the selection and order of demonstration examples, and other factors~\citep{topic, liu2024lets}. Additionally, optimizing the conciseness of demonstration examples and improving the computational efficiency of ICL are critical areas of ongoing research~\citep{liu2024incontext}. 
Furthermore, despite preliminary explanations~\citep{dai2022iclft, jiang2023latent}, the underlying working mechanism of ICL remains unclear and requires further investigation.


With the rapid growth of studies in ICL, our survey aims to sensitize the community toward the current progress.
In the following sections, we delve into an in-depth discussion of related studies, and we summarize the taxonomy in Figure~\ref{taxo_of_icl} and the key findings in Appendix~\ref{app:takeaway}. 
We highlight the challenges and potential directions and hope our work provide a useful roadmap for beginners interested in this area and shed light on future research.



\section{Definition and Formulation}
\label{sec:formulation}
Following \citet{gpt3}, we here provide a formal definition of in-context learning:

\begin{quote}
\textsl{In-context learning is a paradigm that allows language models to learn tasks given only a few examples in the form of demonstration.}
\end{quote}


Formally, given a query input text $x$ and a set of candidate answers $Y = \{y_1, \ldots, y_m\}$, a pretrained language model $\mathcal{M}$ takes the candidate answer with the maximum score as the prediction,\footnote{$Y$ could be class labels or a set of free-text phrases.} conditioned a demonstration set $C$.
$C$ contains an optional task instruction $I$ and $k$ demonstration examples, thus $C = \{ I, s(x_1, y_1), \ldots, s(x_k, y_k) \}$ or $C = \{ s^{\prime}(x_1, y_1, I), \ldots, s^{\prime}(x_k, y_k, I) \}$, where $s^{\prime}(x_i, y_i, I)$ is an example written in natural language according to the task.
Depending on whether $k$ and the demonstration examples belong to the same task, it can be categorized as task-specific ICL and cross-task ICL. In the latter, different examples have their own instructions.
The likelihood of a candidate answer $y_j$ comes from a scoring function $f$ on the whole input sequence:
\begin{equation}
    P( y_j \mid x) \triangleq
    f_\mathcal{M} ( y_j,  C, x)
\end{equation}
The final predicted label $\hat y$ is the candidate answer with the highest probability:
\begin{equation}
    \hat y = \arg\max_{y_j \in Y } P(y_j \mid x). 
\end{equation}

According to the definition, we can see that ICL differs from related concepts as follows: (1) \textit{Prompt Learning}: prompts can be discrete templates or soft parameters that encourage the model to predict the desired output. ICL can be regarded as a subclass of prompt tuning where the demonstration examples are part of the prompt. \citet{liu2021pre} made a thorough survey on prompt learning, but ICL was not included in their study. (2) \textit{Few-shot Learning}: few-shot learning is a general machine learning approach that involves adapting model parameters to perform a task with a limited number of supervised examples~\cite{wang2019few}. In contrast, ICL does not require parameter updates and is directly performed on pretrained LLMs.

\section{Model Training}
\label{sec:training}
Although LLMs have demonstrated promising ICL capability directly, many studies revealed that these ICL capabilities can be further enhanced through specialized training before inference~\cite{selfsupericl, picl, Shi2023iclm}.



\subsection{Pretraining}
\label{sec:pretraining}
One straightforward direction to boost the ICL capability of LLMs is through pretraining or continual pretraining.
For instance, \citet{picl} and \citet{Shi2023iclm} proposed to reorganize pretraining corpora by aggregating related contexts, making models learn to reason across prior demonstrations. Differently, \citet{Li2023mend} introduced a meta-distillation pretraining process, which allows LLMs to reason with distilled demonstration vectors, thereby enhancing ICL efficiency without compromising its effectiveness.

\begin{figure}
    \centering
    \includegraphics[width=0.92\columnwidth]{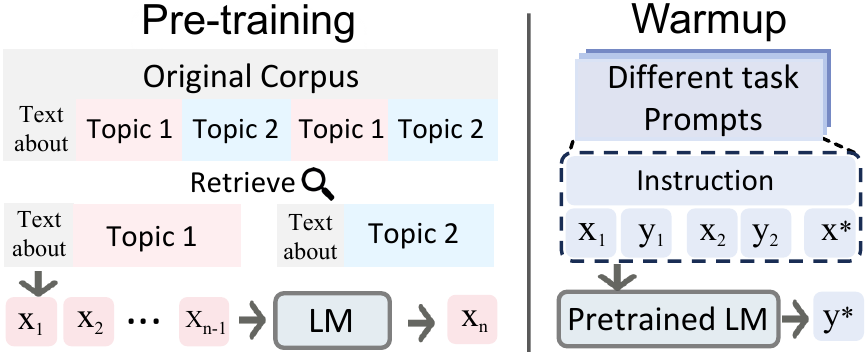}
    \caption{Illustration of model training methods to enhance ICL capabilities through two different stages: pretraining and warmup.}
    \label{fig:train_method}
\end{figure}

\subsection{Warmup}
\label{sec:warmup}
Another way to enhance ICL ability is adding a continual training stage between pretraining and ICL inference, which we call model warmup for short. 
Warmup is an optional procedure for ICL, which adjusts LLMs before inference by modifying or adding parameters.

As most pretraining data are not tailored for ICL~\citep{selfsupericl}, researchers have introduced various warmup strategies to bridge the gap between pretraining and ICL inference. Both \citet{metaicl} and \citet{natural} proposed to continually finetune LLMs on a broad range of tasks with multiple demonstration examples, which boosts ICL abilities.
To encourage the model to learn input-label mappings from the context, \citet{symboltuning} proposed symbol tuning, which substitutes natural language labels (e.g., ``positive/negative sentiment'') with arbitrary symbols (e.g., ``foo/bar''). \citet{selfsupericl} proposed a self-supervised method to align raw text with ICL formats in downstream tasks. Besides, multiple studies have indicated the potential value of instructions~\cite{mishra2021cross, flan}. Tuning the 137B LaMDA-PT~\cite{lamda} on over 60 datasets verbalized via natural language instruction templates, FLAN~\cite{flan} improves the ability of LLMs to follow instructions, boosting both the zero-shot and few-shot ICL performance. \citet{chung} and \citet{natural} proposed to further scale up instruction tuning with more than 1000+ task instructions.

\section{Prompt Designing}
\label{sec:prompt_tuning}
\label{sec:demo}
\begin{table*}[t!]
    \centering
    \vspace{-1.0cm}
    \setlength{\tabcolsep}{2pt}
    \small
    \resizebox{\linewidth}{!}{\begin{tabular}{@{}lcccc@{}}  
    \toprule 
    \bf Category & \bf Methods &  \bf Demonstration Acquisition & \bf LLMs & \bf Features \\
    \midrule 
    \multirow{6}{*}{\shortstack{ Demonstration \\ Selection}}  
    & KATE~\citep{liu2022close} & Human design & GPT-3 & KNN Selection \\
    & MI~\citep{sorensen2022information} &Human design & GPT-3  & Mutual Information \\
    & EPR~\citep{rubin2022learning}  & Human design & GPT-\{J, 3\}/CodeX & Score-based Retrieval \\ 
    & IDS~\citep{qin2023incontext}   & Human design  & GPT-3.5   & Iterative Selection \\
    & AdaICL~\citep{mavromatis2023examples} &Human design & GPT-\{J, Neo\}  & Selective Demonstration \\
    & UDR~\citep{udr} &Human design & GPT-Neo-2.7B  & Unified Retrieval \\\midrule
    \multirow{4}{*}{\shortstack{ Demonstration \\ Reformatting}}
    & SG-ICL~\cite{kim2022self} & LM generated & GPT-J & Auto Demonstration Generation\\
    & AutoICL~\citep{yang2023autoicl} &LM generated & GPT-3.5-Turbo-0301  & Reasoning Path Generation \\
    & MSP~\citep{yang2023demonstration} &Human design & GPT series & Adjusting Demonstration Weight \\ 
    & ICV~\citep{liu2024incontext} &Human design & Falcon-7b / Llama-7b & Demonstration Embedding \\\midrule
    \multirow{2}{*}{\shortstack{Demonstration \\ Ordering}}
    & GlobalE \& LocalE~\citep{lu2022order} &  Human design & GPT-\{2, 3\} & Best Order Selection\\ 
    & ICCL~\citep{liu2024lets} &Human design & Llama2/Mixtral/Qwen & Ordering from Simple to Complex \\ 
    \bottomrule
    \end{tabular}}
    \caption{Summary of representative demonstration designing methods.}
    \label{tab:promptmethods}
\end{table*}

In this section, we focus on the principles of ICL during inference, including demonstration organization~(\S \ref{sec:demonstration_org})  and instruction formatting~(\S \ref{sec:instruction}) . 

\subsection{Demonstration Organization}
\label{sec:demonstration_org}
Many studies have shown that the performance of ICL strongly relies on the demonstration surface, including the selection, formatting, and ordering of demonstration examples~\citep{calibrate, lu2022order}. 
In this subsection, we survey demonstration organization strategies and classify them into three categories, as shown in Table~\ref{tab:promptmethods}.

\subsubsection{Demonstration Selection}
\label{sec:select}

Demonstrations selection aims to answer a fundamental question: \textsl{Which samples are good examples for ICL?} We categorize the related studies into two approaches: unsupervised methods based on predefined metrics and supervised methods. 

\paragraph{Unsupervised Method} A straightforward approach to selecting ICL examples is to choose the nearest neighbors of input instances based on their similarities~\cite{liu2022close, tanwar2023multilingual, qin2023incontext}. Distance metrics, such as L2 distance or cosine similarity based on sentence embeddings, are commonly used for this purpose. For example, \citet{liu2022close} proposed KATE, the first $k$NN-based unsupervised retriever for selecting in-context examples. Similarly, $k$-NN cross-lingual demonstrations can be retrieved for multi-lingual ICL to strengthen source-target language alignment~\citep{tanwar2023multilingual}. \citet{su2022selective} proposed to combine graphs and confidence scores to select diverse and representative examples. In addition to distance metrics, mutual information~\cite{sorensen2022information} and perplexity~\cite{gonen2022demystifying} have proven valuable for prompt selection without labeled examples or specific LLMs. Furthermore, using output scores 
of LLMs as unsupervised metrics has shown effectiveness in demonstration selection~\citep{Wu2022SelfadaptiveIL, nguyen2023influence, li2023supporting}. Particularly, \citet{Wu2022SelfadaptiveIL} selected the best subset permutation of $k$NN examples based on the code length for data transmission to compress label $y$ given $x$ and $C$. \citet{li2023supporting} used infoscore, i.e., the average of $P(y|x_i,y_i,x)   P(y|x)$ for all $(x,y)$ pairs in a validation set with a diversity regularization. 

\paragraph{Supervised Method} Though off-the-shelf retrievers offer convenient services for extensive NLP tasks, they are heuristic and sub-optimal due to the lack of task-specific supervision. To address this issue, numerous supervised methods have been developed~\cite{rubin2022learning, ye2023compositional, topic, zhang2022active}. EPR~\cite{rubin2022learning} introduced a two-stage method to train a dense retriever for demonstration selection. For a specific input, it first utilized unsupervised methods (e.g., BM25) to recall similar examples as candidates and then used this data to build a supervised dense retriever. Following EPR, \citet{udr} adopted a unified demonstration retriever to select demonstrations across different tasks. Unlike prior work that retrieves individual demonstrations, \citet{ye2023compositional} proposed retrieving entire demonstration sets to model inter-relationships between examples. Additionally, \citet{mavromatis2023examples} introduced AdaICL, a model-adaptive method that employs LLM to predict the unlabeled data set, generating an uncertainty score for each instance.

Based on prompt tuning, \citet{topic} viewed LLMs as topic models that can infer concepts $\theta$ from a few demonstrations and generate tokens based on these concepts. They represent latent concepts with task-related concept tokens, which are learned to maximize $P(y|x,\theta)$. Demonstrations are selected based on their likelihood to infer the concept variable using $P(\theta|x,y)$. Additionally, reinforcement learning was introduced by \citet{zhang2022active} for example selection. They formulated demonstration selection as a Markov decision process~\cite{bellman1957markovian} and selected demonstrations via Q-learning. The action is choosing an example, and the reward is defined as the accuracy of a labeled validation set.

\begin{table}[t]                
\centering      
\setlength{\tabcolsep}{2pt}  
{  \fontsize{9pt}{11pt}\selectfont          
\begin{tabular}{lccccccc}      
\toprule             
\bf Model & \bf Method & \bf SST5 &  \bf SST2 & \bf CQA  & \bf SNLI & \bf News & \bf Avg \\       
\midrule             
\multirow{3}{*}{GPT2}          
& topk & 40.1 & 74.9 & 30.2 & 39.7&62.7 & 49.5\\            
& votek & 32.4 & 51.0 & 29.8 & 35.8& 25.5 & 34.9 \\            
& mdl & \textbf{43.3} & \textbf{86.7} & \textbf{32.7} & \textbf{41.4}& 
\textbf{68.0} & \textbf{54.4}\\       
\midrule             
\multirow{3}{*}{GPT-J}              
& topk & \textbf{46.9} & 84.6 & 58.4 & \textbf{60.7} & \textbf{69.1} & \textbf{63.9} \\            
& votek & 33.8 & 87.3 & 63.4 & 43.1& 25.3 & 50.6\\            
& mdl & 37.6 & \textbf{87.9} & \textbf{64.1} & 59.8  & 68.2 &63.5\\        
\midrule              
\multirow{3}{*}{Qwen2}        
& topk & 54.1 & 83.3 & 76.3 & \textbf{68.2} &64.9 &  \textbf{69.4}\\            
& votek & \textbf{55.3} & \textbf{86.9} & 76.1 &51.6& \bf 65.3 & 67.0\\            
& mdl & 54.6 & 86.1 & \textbf{77.1} &65.0& 63.2 &69.2\\            
\midrule             
\multirow{3}{*}{Llama3}              
& topk & 53.0 & \textbf{90.3} & 76.1 & \textbf{64.0} & 74.0 &  \textbf{71.5}\\            
& votek & 54.9 & 88.9 & 72.6 & 57.7& \textbf{78.3} & 70.5\\            
& mdl & \textbf{54.4} & 89.1 & \textbf{76.5} & 59.9 & 74.6 &70.9 \\               
\bottomrule                
\end{tabular}}                   
\caption{Fair comparison of demonstration selection methods. CQA and News are abbreviations of Commonsense QA and AG News, respectively. The best results are \textbf{bolded}. Our experiments on topk~\citep{liu2022close}, votek~\citep{su2022selective}, mdl~\citep{Wu2022SelfadaptiveIL} 
show that the effectiveness of ICL example selection methods are model-dependent. On GPT-2, the mdl method performs the best, while on the other three models,
topk performs the best.
}    
\label{tab:experiment_design_centered_model_names}        
\end{table}

In order to have a more intuitive comparison of the performance of several unsupervised methods, we select topk~\citep{liu2022close}, votek~\citep{su2022selective}, mdl~\citep{Wu2022SelfadaptiveIL} 
to conduct experiments. The result is shown in Table 2. The details of the experiment can be found in Appendix \ref{app:experiment}.





\subsubsection{Demonstration Reformatting}
\label{sec:reformatting}
In addition to directly selecting examples from training data, another research trend involves utilizing LLMs to reformat the representation of existing demonstrations~\cite{kim2022self, yang2023autoicl, hao2022structured, yang2023demonstration, liu2024incontext, li2024featureadaptive}. For instance, \citet{kim2022self} proposed generating demonstrations directly from LLMs to reduce the reliance on external demonstration data. Structured Prompting \citep{hao2022structured} proposed to encode demonstration examples separately with special positional embeddings, which are then provided to the test examples using a rescaled attention mechanism. Diverging from these methods, other approaches focus on modifying the latent representation of demonstrations~\citep{liu2024incontext, li2024featureadaptive}. Specifically, \citet{liu2024incontext} developed In-Context Vectors (ICVs) derived from the latent embeddings of demonstration examples in LLMs. These ICVs are used during inference to adjust the latent states of the LLM, thereby enhancing the model's ability to follow the demonstrations more effectively.

\subsubsection{Demonstration Ordering}
\label{sec:order}
Ordering the selected demonstration examples is also an important aspect of demonstration organization. \citet{lu2022order} have proven that order sensitivity is a common problem and always exists for various models. To handle this problem, previous studies have proposed several training-free methods for sorting demonstration examples. Particularly, \citet{liu2022close} arranged examples based on their proximity to the input, positioning the closest example as the rightmost demonstration. \citet{lu2022order} introduced global and local entropy metrics, finding a positive correlation between these metrics and the ICL performance. Consequently, they utilized the entropy metric to determine the optimal demonstration ordering. Additionally, ICCL~\citep{liu2024lets} suggested ranking demonstrations from simple to complex, thereby gradually increasing the complexity of demonstration examples during the inference process.
\subsection{Instruction Formatting}
\label{sec:instruction}
A common way to format demonstrations is concatenating examples $(x_1, y_1), \ldots, (x_k, y_k)$ with a template $\mathcal{T}$ directly. However, in some tasks that need complex reasoning (e.g., math word problems and commonsense reasoning), it is not easy to learn the mapping from $x_i$ to $y_i$ with only $k$ demonstrations. Although template engineering has been studied in prompting~\citep{liu2021pre},  some researchers aim to design a better format of demonstrations for ICL by describing tasks with the instruction $I$. \citet{induct} found that given several demonstration examples, LLMs can generate task instructions themselves. Considering the generation abilities of LLMs, \citet{zhou2022large} proposed an Automatic Prompt Engineer for automatic instruction generation and selection.
To further improve the quality of the automatically generated instructions, several strategies have proposed using LLMs to bootstrap off its own generations~\citep{wang2022self, chen2024grimoire}. 
Additionally, chain-of-thought (CoT)~\cite{cot} introduces intermediate reasoning steps between inputs and outputs to enhance problem-solving and comprehension. Recent advancements have also emphasized the process of enhancing step-by-step reasoning in models~\citep{autocot, wang2022iteratively, least}.
\subsection{Scoring Function}
\label{sec:scoring}

\begin{table}[t!]
    \centering
    \resizebox{\linewidth}{!}{
    \begin{tabular}{@{}l|c|ccc@{}}
    \toprule 
      \bf Method & \bf Target & \bf Efficiency & \bf Coverage & \bf Stability \\
    \midrule 
       Direct  & $ \mathcal{M} ( y_j \mid C, x) $ & +++  &  + & + \\ 
       PPL &  $\text{PPL} (S_j) $&  + & +++ &  +\\ 
       Channel& $\mathcal{M} (x \mid C, y_j)$ & + & +  & ++ \\ 
    \bottomrule
    \end{tabular}
    }
    \caption{Summary of different scoring functions. Coverage refers to task coverage. The qualitative results for `Efficiency' and `Stability' metrics are elaborated in  Table~\ref{tab:appendix_efficiency} and Table~\ref{tab:appendix_stability}, respectively.}
    \label{tab:score_func}
\end{table}

The scoring function determines how to transform the predictions of a language model into an estimation of the likelihood of a specific answer. The Direct method uses the conditional probability of candidate answers represented by tokens in the model's vocabulary \citep{gpt3}. The answer with the highest probability is selected as the final answer, but this method restricts template design by requiring answer tokens to be at the end of input sequences.
Perplexity (PPL) is another commonly used metric that computes the sentence perplexity of the entire input sequence \( S_j = \{ C, s(x, y_j, I) \} \), which includes tokens from demonstration examples \( C \), the input query \( x \), and the candidate label \( y_j \). PPL evaluates the probability of the sentence, eliminating token position limitations but requiring additional computation time. \citet{min2022noisy} proposed using channel models (Channel) to compute the conditional probability in reverse, estimating the likelihood of the input query given the label. This approach requires language models to generate every token in the input, potentially boosting performance under imbalanced training data. We summarize all three scoring functions in Table~\ref{tab:score_func}.
Note that in Table~\ref{tab:score_func}, `Efficiency' refers to the language model inference latency; `Coverage' reflects whether the method utilizes the output probability of the local or all token positions in the input sequence; and `Stability' indicates whether the in-context learning ability is easily affected by changes in the demonstration examples.

\section{Analysis}
\label{sec:analysis}
\begin{figure*}
    \centering
    \vspace{-1.0cm}
    \includegraphics[width=0.9\textwidth]{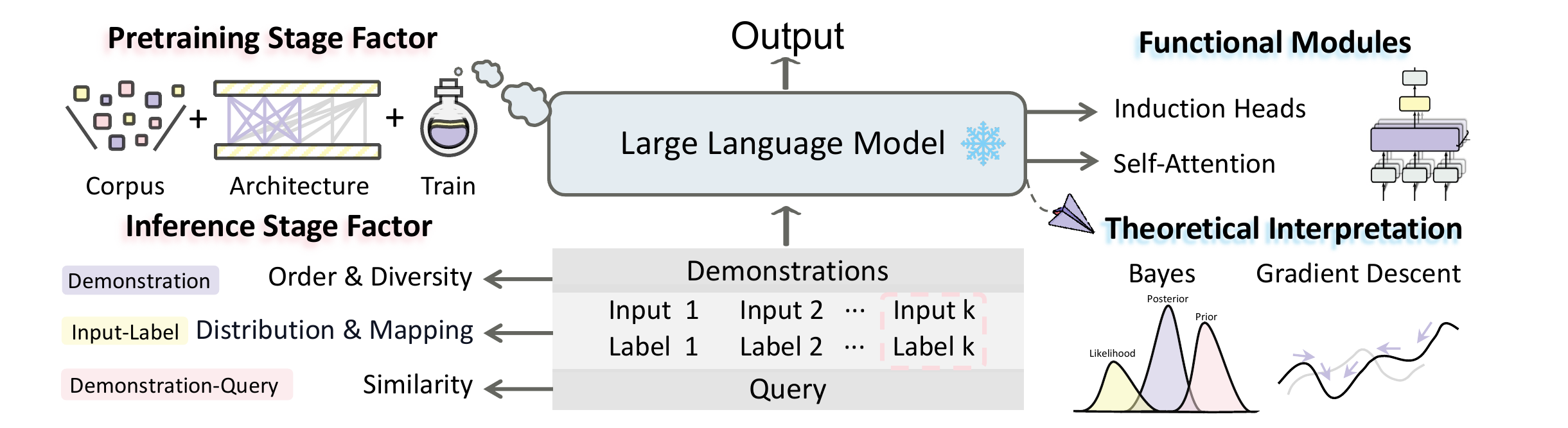}
    \caption{Summary of factors that have a relatively strong correlation to ICL performance and different perspectives to explain why ICL works.}
    \label{fig:factor}
\end{figure*}

To understand ICL, recent studies attempt to investigate what influence ICL performance~\cite{shin2022corpora, ground_truth, Kossen2023InContextLL} and why ICL works~\cite{dai2022iclft, Irie2022TheDF}. 
In this section, we present a detailed elaboration of influencing factors~(\S \ref{sec:inf_factors}) and learning mechanisms~(\S \ref{sec:mech}) of ICL, as illustrated in Figure~\ref{fig:factor}.
\subsection{Influencing Factors}
\label{sec:inf_factors}
We discuss relevant research addressing \textsl{what influences ICL performance}, including factors both in the pretraining stage and in the inference stage.

\subsubsection{Pretraining Stage} 
\label{sec:inf_factors_pre}
We first introduce factors that influence the pretraining stage. The diversity of pretraining corpora significantly impacts ICL performance~\cite{shin2022corpora, Yadlowsky2023PretrainingDM, Raventos2023PretrainingTD}. 
In particular, \citet{shin2022corpora} found that the source domain is more important than the corpus size, suggesting that combining multiple corpora may lead to the emergence of ICL ability. Similarly, \citet{Raventos2023PretrainingTD} empirically identified a task diversity threshold beyond which LLMs exhibit strong ICL capabilities in unseen tasks.
Another line of research investigates the impact of data distribution on ICL~\cite{distribution, wies2024learnability}. For instance, \citet{distribution} demonstrated that ICL capability emerges when the training data exhibits specific distributional properties, such as burstiness, wherein items appear in clusters rather than being uniformly distributed over time.

Beyond these works, several studies have investigated the impact of model architecture and training process on ICL performance~\cite{wei2022emergent, gpt3, Ding2023CausalLMIN}. \citet{wei2022emergent} investigated the emergent abilities of many large-scale models on multiple tasks. They suggested that a pretrained model acquires some emergent ICL abilities when it reaches a large scale of pretraining steps or model parameters. \citet{Ding2023CausalLMIN} pointed out that the in-context samples should attend to each other during inference, indicating that current causal LLMs may lead to suboptimal ICL performance.

\subsubsection{Inference Stage}
\label{sec:inf_factors_infe}
During inference, there are also multiple properties of demonstration examples that influence ICL performance. \citet{min2022rethinking} proved that input-label settings such as the pairing format, the exposure of label space, and the input distribution contribute substantially to ICL performance. 
However, contrary to the conclusion in \citet{min2022rethinking} that
input-label mapping matters little to ICL, latter studies showed that the accurate mapping influence ICL performance significantly~\cite{ground_truth, Pan2023WhatIL, Tang2023LargeLM}. \citet{llm_icl_differently} further pointed that flipped or semantically-unrelated input-label mapping also can be learned.
%

From the perspective of demonstration construction, recent literature focuses on the diversity and simplicity of demonstrations~\cite{compositional_generalization}, the order of samples~\cite{lu2022order, Zhang2022ActiveES, Liu2023LostIT}, and the similarity between demonstrations and queries~\cite{liu2022close}. For example, \citet{liu2022close} found that demonstration samples with embeddings closer to those of the query samples typically yield better performance than those with more distant embeddings.
Notably, despite efforts to refine demonstrations to optimize the performance, there still remain clear feature biases during ICL inference~\cite{inductive_bias}. Overcoming strong prior biases and ensuring the model gives equal weight to all contextual information remain challenges~\cite{Kossen2023InContextLL}.


\subsection{Learning Mechanism}
\label{sec:mech}

From a learning mechanism perspective, we delve into the research addressing why ICL is effective.



\subsubsection{Functional Modules} 
\label{sec:mech_fun}
The ICL capability is intimately connected to specific functional modules within Transformers.
As one of the core components, the attention module is a focal point in the study of ICL mechanism~\cite{olsson2022induction, Bietti2023BirthOA,dai2022iclft, Irie2022TheDF, icl_weight_shifting,Gao2023InContextLF,zhang2023and}. Particularly, \citet{olsson2022induction} identified specific attention heads, referred to as ``induction heads'', that can replicate previous patterns for next-token prediction, thus progressively developing ICL capabilities.
Additionally, \citet{label_anchor} focused on the information flow in Transformers and found that during the ICL process, demonstration label words serve as anchors, which aggregate and distribute key information for the final prediction.


\subsubsection{Theoretical Interpretation}
\label{sec:mech_theo}
In this subsection, we introduce the theoretical interpretations of ICL from different views.

\paragraph{Bayesian View} In the Bayesian framework, ICL is explained as implicit Bayesian inference, where models perform ICL by identifying a shared latent concept among examples~\cite{bayesian,wies2024learnability,ahuja2023context,jiang2023latent,topic}. Additional perspectives suggest that LLMs encode the Bayesian Model Averaging algorithm via the attention mechanism~\cite{zhang2023and}. As the number of in-context examples increases, implicit Bayesian inference becomes analogous to kernel regression~\cite{Han2023ExplainingEI}.

\paragraph{Gradient Descent View} Gradient descent offers another valuable lens for understanding ICL. ~\citet{dai2022iclft} identified a dual form between Transformer attention and gradient descent, finding that GPT-based ICL behaves similarly to explicit fine-tuning from multiple perspectives. Other studies have attempted to establish connections between ICL and gradient descent in simplified regression settings~\cite{icl_gd, Ahn2023TransformersLT, mahankali2023one, icl_weight_shifting}. For instance, \citet{icl_gd} showed that linear attention-only Transformers with manually constructed parameters are closely related to models learned by gradient descent. \citet{icl_weight_shifting} found that self-attention-only Transformers exhibit similarities with models trained via gradient descent. However, the simplified settings used in these studies have led to debates about the direct applicability of these connections in real-world contexts~\cite{Shen2023RevisitingTH}. \citet{fu2023transformers} argued that Transformers perform ICL on linear regression using higher-order optimization techniques rather than gradient descent.

\paragraph{Other Views} Beyond connecting ICL with a single algorithm, researchers have analyzed it from various perspectives, including ability decoupling, algorithmic learning, and information theory. \citet{tr_and_tl} decoupled ICL capabilities into task recognition ability and task learning ability, each manifesting under different conditions. Another typical theory abstracts ICL as an algorithmic learning problem~\cite{akyurek2022algorithm, garg2022linear, trm_as_alg, Bai2023TransformersAS}, where Transformers dynamically select algorithms, such as gradient descent and ridge regression, tailored to different ICL instances. Moreover, \citet{implicit_structure_induction} utilized information theory to show an error bound for ICL under linguistically motivated assumptions, explaining how next-token prediction can bring about the ICL ability.

These analytical studies have taken an essential step to explain ICL. However, most of them focused on simple tasks and small models. Extending analysis on extensive tasks and large models may be the next step to be considered. 


\section{Application}
\label{sec:application}
\label{app}
Given its user-friendly interface and lightweight prompting method, ICL has broad applications on traditional NLP tasks~\cite{kim2022self,metaicl,zhu2023multilingual}.
Particularly, by using demonstrations that explicitly guide the reasoning process, ICL manifests remarkable effects on tasks requiring complex reasoning~\cite{cot,li2023code,teachalgo} and compositional generalization~\cite{least}. 



We explore several emerging and prevalent applications of ICL, including data engineering, model augmentation, and knowledge updating.
\textbf{1) Data Engineering:} Unlike traditional methods such as human annotation and noisy automatic annotation, ICL generates relatively high-quality data at a lower cost, leading to improved performance.~\cite{want, khorashadizadeh2023exploring, annotation}.
\textbf{2) Model Augmentation:} The context-flexible nature of ICL shows promise in model augmentation. It can enhance retrieval-augmented methods by prepending grounding documents to the input~\cite{ram2023context}. Additionally, ICL for retrieval demonstrates potential in steering models toward safer outputs~\cite{dpicl, meade2023using}.
\textbf{3) Knowledge Updating:} LLMs often contain outdated or incorrect knowledge~\cite{NEURIPS2023_5f0a4cd2}. ICL has demonstrated efficacy in revising such knowledge through carefully crafted demonstrations, yielding higher success rates compared to gradient-based methods~\cite{editingfact}.

As mentioned above, ICL has yielded significant benefits on both traditional and emergent NLP applications. 
The tremendous success of ICL in NLP has inspired researchers to explore its potential in various modalities beyond text (elaborated in Appendix~\ref{app:vision}), including vision ~\cite{bar2022visual_icl,wang2023imagesPainter}, vision-language~\cite{tsimpoukelli2021frozen, alayrac2022flamingo}, as well as speech applications~\cite{wang2023neural,zhang2023valle-x}.

\section{Challenges and Future Directions}
\label{sec:challege_future}
In this section, we review existing challenges and discuss future directions for ICL.

\paragraph{Efficiency and Scalability}
The use of demonstrations in ICL introduces two challenges: (1) higher computational costs with an increasing number of demonstrations (\textit{efficiency}), and (2) fewer learnable samples due to the maximum input length of LLMs (\textit{scalability}). Prior research has attempted to mitigate these issues by distilling lengthy demonstrations into compact vectors~\cite{Li2024ImplicitIL,Li2023mend} or expediting LLM inference times~\cite{Liu2023dejavu}. However, these methods often involve a trade-off in performance or necessitate access to model parameters, which is impractical for closed-source models like ChatGPT and Claude~\cite{Zhou2023Efficient}. Thus, enhancing the scalability and efficiency of ICL with more demonstrations remains a significant challenge.


\paragraph{Generalization}
ICL heavily relies on high-quality demonstrations selected from annotated examples, which are often scarce in low-resource languages and tasks. This scarcity poses a challenge to the generalization ability of ICL~\cite{He2024SelfDemos}. Given that there is a substantial discrepancy in the availability of annotated high-resource data and low-resource data, the potential to leverage high-resource data to address low-resource tasks is highly appealing~\cite{Chatterjee2024LanguageMC, tanwar2023multilingual}.



\paragraph{Long-context ICL}


Recent advances in context-extended LLMs have spurred research into the impact of ICL when using an increasing number of demonstration examples~\cite{agarwal2024manyshot,Bertsch2024InContextLW}. However, researchers have found that increasing the number of demonstrations does not necessarily enhance performance and may even be detrimental. These performance declines indicate a need for further investigation. Additionally, \citet{Li2024LongcontextLS} developed LongICLBench, which includes diverse extreme-label classification tasks, revealing further weaknesses of LLMs in comprehending extended demonstrations.

\section{Conclusion}
\label{sec:conclusion}
In this paper, we comprehensively review the existing literature on ICL, examining advanced techniques, conducting analytical studies, discussing relevant applications, and identifying critical challenges and potential directions for future research. To our knowledge, this is the first comprehensive survey dedicated to ICL. We aim to highlight the current state of research in ICL and provide insights to guide future work in this promising area.

\section*{Limitations}
This paper offers a comprehensive examination and summary of current methodologies and analyses in the area of In-Context Learning (ICL). However, given the extensive body of related work, particularly in demonstration design and the principle analysis of ICL, we may have overlooked some equally valuable contributions. Additionally, we outline several future directions for research in ICL, including long-context ICL, efficiency and scalability in ICL, etc. We plan to leave these aspects for future work. 
Furthermore, many papers covered by this survey did not utilize the most up-to-date models while running experiments. We advocate for more thorough and up-to-date research to provide actionable insights for practitioners.

\bibliography{custom}

\appendix

\section{Takeaway}
\label{takeaway}
\label{app:takeaway}
Through a comprehensive literature review of ICL, we have discovered takeaways across several domains. These include training, demonstration design, scoring functions, analysis, and ICL applications that go beyond text.

\subsection{Training}
To further enhanced ICL capabilities, methods propose to train the LLMs in the stage of pre-training and warmup before ICL inference.

 \textbf{$\Diamond$ Takeaway}: 
(1) The key idea of training before inference is to bridge the gap between pretraining and downstream ICL formats by introducing objectives close to in-context learning. Warmup is optional for ICL as many pretrained LLMs have manifested the ICL ability.  
(2) Compared to in-context finetuning involving demonstration, instruction finetuning without a few examples as demonstration is simpler and more popular. All these warmup methods improve the ICL capability by updating the model parameters, which implies that the ICL capability of the original LLMs has great potential for improvement. Therefore, although ICL does not strictly require model warmup, we recommend adding a warmup stage before ICL inference.
(3) 
The performance advancement made by warmup encounters a plateau when increasingly scaling up the training data, indicating that LLMs only need a small amount of data to adapt to learn from the context during warmup.

\subsection{Demonstration Organization}

The performance of ICL strongly relies on the demonstration surface, including the selection, formatting, and ordering of demonstration examples.

 \textbf{$\Diamond$ Takeaway}: 
 (1) Demonstration selection strategies improve the ICL performance, but most of them are instance level. Since ICL is mainly evaluated under few-shot settings, the corpus-level selection strategy is more important yet underexplored. 
 (2) The output score or probability distribution of LLMs plays an important role in instance selecting. 
 (3) For k demonstrations, the size of search space of permutations is k!. How to find the best orders efficiently or how to approximate the optimal ranking better is also a challenging question. 
(4) Adding chain-of-thoughts can effectively decompose complex reasoning tasks into intermediate reasoning steps. During inference, multi-stage demonstration designing strategies are applied to generate CoTs better. How to improve the CoT prompting ability of LLMs is also worth exploring. 
(5) In addition to human-written demonstrations, the generative nature of LLMs can be utilized in demonstration designing. LLMs can generate instructions, demonstrations, probing sets, chain-of-thoughts, and so on. By using LLM-generated demonstrations, ICL can largely get rid of human efforts on writing templates.

\subsection{Scoring Function}
The scoring function determines how to transform the predictions of a language model into an estimation of the likelihood of a specific answer. The answer with the highest probability is selected as the final answer.

 \textbf{$\Diamond$ Takeaway}: 
(1) Although directly adopting the conditional probability of candidate answers is efficient, this method still poses some restrictions on the template design. Perplexity is also a simple and widely scoring function. This method has universal applications, including both classification tasks and generation tasks. However, both methods are still sensitive to demonstration surface, while Channel is a remedy that especially works under imbalanced data regimes. 
(2) Existing scoring functions all compute a score straightforwardly from the conditional probability of LLMs. There is limited research on calibrating the bias or mitigating the sensitivity via scoring strategies. 

\subsection{Analysis}
Numerous analytical studies investigate influencing factors of ICL during both the pretraining and inference stages, and attempt to figure out the learning mechanisms of ICL from the perspective of functional modules and theoretical interpretation.

 \textbf{$\Diamond$ Takeaway}: 
(1)
Knowing and considering why ICL works and what factors may influence can help us improve the ICL performance.
(2) 
Although some analytical studies have taken a preliminary step to explain ICL, most of them are limited to simple tasks and small models. 
Extending analysis on extensive tasks and large models may be the next step to be considered. 
(3) Among existing work, explaining ICL with gradient descent seems to be a reasonable, general, and promising direction for future research. 
If we build clear connections between ICL and gradient-descent-based learning, we can borrow ideas from the history of traditional deep learning to improve ICL.

\subsection{In-context Learning Beyond Text}
The tremendous success of ICL in NLP has inspired researchers to explore in-context learning in different modalities beyond natural language with promising results.

\textbf{$\Diamond$ Takeaway}: 
(1) Properly formatted data (e.g., interleaved image-text datasets for vision-language tasks) and architecture designs are key factors for activating the potential of in-context learning. Exploring it in a more complex structured space such as for graph data is challenging and promising~\citep{huang2023graph_icl}.
(2) Findings in textual in-context learning demonstration design and selection cannot be trivially transferred to other modalities. Domain-specific investigation is required to fully leverage the potential of in-context learning in various modalities.


\section{Experimental Detail}
\label{app:experiment}
In the experiment, we utilize 8 demonstrations and test on gpt2~\citep{gpt2}, gptj~\citep{gpt-j}, LLaMA3-8B-Instruct\citep{llama3} and Qwen2-7B-Instruct~\citep{qwen}. All experiments are executed on a single NVIDIA A100 (80G). For datasets we choose sst2~\citep{sst2}, sst5~\citep{sst5}, commonsense\_qa~\citep{commonsenseqa}, ag\_news~\citep{agnews} and snli~\citep{snli}. For the last two datasets, we only select 1000 data from the training set for retrieval and the first 1000 data from the test set for testing. During the inference phase, a PPL-based approach is employed. The entire code framework is built upon OpenICL~\citep{openicl}, for which we extend our gratitude to the authors.

Table~\ref{tab:appendix_efficiency} and Table~\ref{tab:appendix_stability} show the quantitative results on the efficiency and stability metrics for different scoring functions in Table~\ref{tab:score_func}.

\begin{table}[t]                
\centering      
\setlength{\tabcolsep}{4pt}  
{  \fontsize{9pt}{11pt}\selectfont          
\begin{tabular}{lccc} 
\toprule             
\bf Model & \bf Direct &  \bf PPL & \bf Channel  \\       
\midrule             
GPT2 &
44.13(1.00) & 114.02(2.58) & 157.70(3.57) \\            
\midrule             
GPT-J &
611.04(1.00) & 1766.82(2.89) & 1793.27(2.93) \\         
\midrule              
Qwen2 & 
745.89(1.00) & 1886.63(2.53) & 1957.97(2.63) \\
\midrule             
Llama3 &
790.46(1.00) & 1935.04(2.45) & 1956.21(2.47) \\
\midrule             
AVG &
\textbf{1.00} & \textbf{2.61} & \textbf{2.90} \\           
\bottomrule                
\end{tabular}}                   
\caption{
The qualitative results of the Efficiency metric in Table~\ref{tab:score_func} which record the language model inference latency (including the time for scoring with different scoring functions, with input data containing 8 in-context examples). The unit is milliseconds (ms). Each cell's parentheses contain the ratio of the latency for the current column model using the current row scoring function to the latency using direct inference. The final calculated average is the average of these ratios.
}    
\label{tab:appendix_efficiency}        
\end{table}

\begin{table}[t]                
\centering      
\setlength{\tabcolsep}{12pt}  
{  \fontsize{9pt}{11pt}\selectfont          
\begin{tabular}{lccc} 
\toprule             
\bf Model & \bf Direct &  \bf PPL & \bf Channel  \\       
\midrule             
GPT2 &
1.12 & 0.85 & 3.18 \\            
\midrule             
GPT-J &
1.00 & 0.77 & 4.06 \\         
\midrule              
Qwen2 & 
0.72 & 0.70 & 2.43 \\
\midrule             
Llama3 &
0.89 & 0.78 & 2.43 \\
\midrule             
AVG &
\textbf{0.93} & \textbf{0.78} & \textbf{3.03} \\           
\bottomrule                
\end{tabular}}                   
\caption{
The qualitative results of the Stability metric in Table~\ref{tab:score_func} which reflect whether the in-context learning ability is easily affected by changes in demonstration examples. We conducted experiments using a test set of size 10k and set up 5 different random seeds. Each time, 8 examples were randomly selected from 5k training examples for the experiments. The table records the variance of performance.
}    
\label{tab:appendix_stability}        
\end{table} 

\section{Evaluation and Resources}
\label{sec:evaluation}

\subsection{Traditional Tasks}
As a general learning paradigm, ICL can be examined on various traditional datasets and benchmarks, e.g., SuperGLUE~\cite{superglue}, SQuAD~\cite{squad}. 
Implementing ICL with 32 randomly sampled examples on SuperGLUE, ~\citet{gpt3} found that GPT-3 can achieve results comparable to state-of-the-art (SOTA) finetuning performance on COPA and ReCoRD, but still falls behind finetuning on most NLU tasks.
~\citet{hao2022structured} showed the potential of scaling up the number of demonstration examples. However, the improvement brought by scaling is very limited. At present, compared to finetuning, there still remains some room for ICL to reach on traditional NLP tasks.


\begin{table}[t!]
    \small
    \setlength{\tabcolsep}{4pt}
    \centering
    \resizebox{\linewidth}{!}{\begin{tabular}{llr}
    \toprule
     \bf Benchmark  & \bf Tasks  & \bf \#Tasks  \\
     \midrule
    \makecell[l]{BIG-Bench \\ \cite{beyond}}   & {\makecell[l]{Mixed tasks}} & 204 \\
    \makecell[l]{BBH \\ \cite{suzgun2022challenging}}   & {\makecell[l]{Unsolved problems}} & 23  \\
    \makecell[l]{PRONTOQA \\ \cite{heuristic}}  & {\makecell[l]{Question answering}} &  1 \\
    \makecell[l]{MGSM \\ \cite{shi2022language}} & {\makecell[l]{Math problems}} & 1  \\
    \makecell[l]{LLMAS \\ \cite{planbench}} & {\makecell[l]{Plan and reasoning tasks}}  & 8\\
    \makecell[l]{OPT-IML Bench \\ \cite{optiml}} & {\makecell[l]{Mixed tasks}} & 2000 \\
    \bottomrule
    \end{tabular}}
    \caption{New challenging evaluation benchmarks for ICL. For short, we use LLMAS to represent LLM Assessment Suite~\cite{planbench}.} 
    \label{tab:dataset}
\end{table}


\subsection{New Challenging Tasks}
In the era of large language models with in-context learning capabilities, researchers are more interested in evaluating the intrinsic capabilities of large language models without downstream task finetuning~\cite{foundation}.

To explore the capability limitations of LLM on various tasks, 
~\citet{beyond} proposed the BIG-Bench~\cite{beyond}, a large benchmark covering  
a large range of tasks, including linguistics, chemistry, biology, social behavior, and beyond. 
The best models have already outperformed the average reported human-rater results on 65\% of the BIG-Bench tasks through ICL~\cite{suzgun2022challenging}. To further explore tasks actually unsolvable by current language models, \citet{suzgun2022challenging} proposed a more challenging ICL benchmark, BIG-Bench Hard (BBH). BBH includes 23 unsolved tasks, constructed by selecting challenging tasks where the state-of-art model performances are far below the human performances. Besides, researchers are searching for inverse scaling tasks,\footnote{\url{https://github.com/inverse-scaling/prize}} that is, tasks where model performance reduces when scaling up the model size. Such tasks also highlight potential issues with the current paradigm of ICL.
To further probe the model generalization ability, ~\citet{optiml} proposed OPT-IML Bench, consisting of 2000 NLP tasks from 8 existing benchmarks, especially benchmark for ICL on held-out categories.

Specifically, a series of studies focus on exploring the reasoning ability of ICL.~\citet{heuristic} generated an example from a synthetic world model
represented in first-order logic and parsed the ICL generations into symbolic proofs for formal analysis. They found that LLMs can make correct individual deduction steps via ICL.
~\citet{shi2022language} constructed the MGSM benchmark to evaluate the chain-of-thought reasoning abilities of LLMs in multilingual settings, finding that LLMs manifest complex reasoning across multiple languages.
To further probe more sophisticated planning and reasoning abilities of LLMs, ~\citet{planbench} provided multiple test cases for evaluating various reasoning abilities on actions and change, where existing ICL methods on LLMs show poor performance.

In addition, ~\citet{tang-etal-2023-context} proposed a benchmark called SAMSum, which is a human-annotated dataset specifically designed for multi-turn dialogue summarization, to evaluate the quality of dialogue summaries generated by LLMs via ICL.

\subsection{Open-source Tools}
Noticing that ICL methods are often implemented differently and evaluated using different LLMs and tasks, \citet{openicl} developed OpenICL, an open-source toolkit enabling flexible and unified ICL assessment. With its adaptable architecture, OpenICL facilitates the combination of distinct components and offers state-of-the-art retrieval and inference techniques to accelerate the integration of ICL into advanced research.

\section{In-Context Learning Beyond Text}
\label{app:vision}


The tremendous success of ICL in NLP  has inspired researchers to explore its potential in different modalities, including visual, vision+language and speech tasks as well. 

\subsection{Visual In-Context Learning}
\begin{figure}
    \centering
    \includegraphics[width=0.49\textwidth]{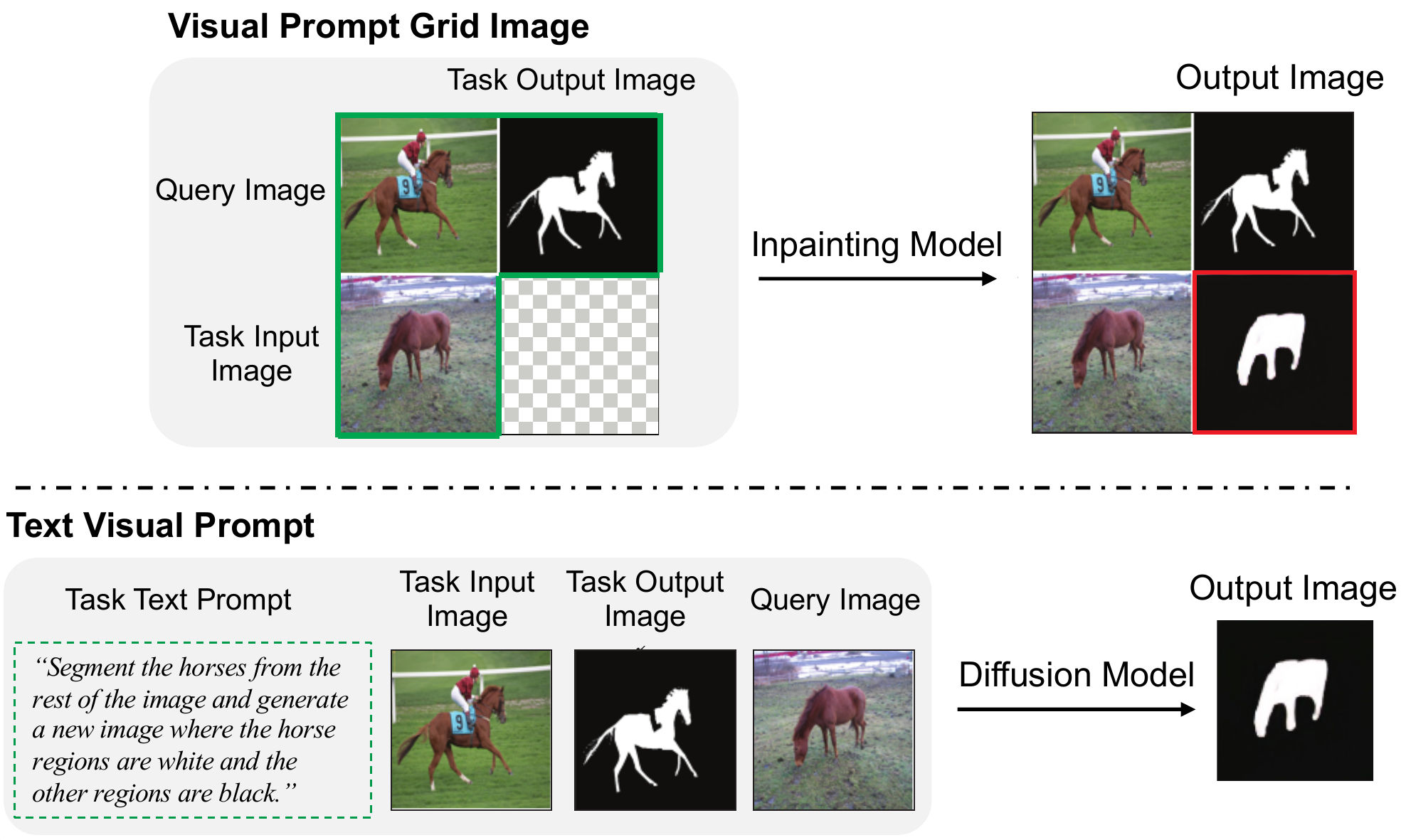}
    \caption{Image-only and textual augmented prompting for visual in-context learning.}
    \label{fig:mm_case}
\end{figure}

Employing masked auto-encoders (MAE) for image patch infilling, the model trained by \citet{bar2022visual_icl} generates consistent output images at inference, demonstrating robust ICL capabilities for tasks like image segmentation. This method is expanded in Painter \citep{wang2023imagesPainter}, which incorporates multiple tasks to develop a generalist model with competitive performance. SegGPT \citep{wang2023seggpt} further builds on this by integrating diverse segmentation tasks and exploring ensemble techniques to enhance example quality. Additionally, \citet{wang2023prompt_diffusion} introduce the Prompt Diffusion model, the first diffusion-based model with ICL abilities, guided by an extra text prompt for more precise image generation, as illustrated in Figure~\ref{fig:mm_case}.

Similar to ICL in NLP, the effectiveness of visual in-context learning greatly depends on the choice of demonstration images, as shown in research by \citep{zhang2023visual_icl_analysis} and \citep{sun2023exploring_visual_icl}. To optimize this, \citet{zhang2023visual_icl_analysis} examine two strategies: using an unsupervised retriever to select the nearest samples with an existing model, and a supervised approach to train a specialized retriever to boost ICL performance. These approaches improve results by ensuring semantic similarity and better alignment in viewpoint, background, and appearance. Beyond retrieval, \citet{sun2023exploring_visual_icl} also investigate a prompt fusion technique to further enhance outcomes.

\subsection{Multi-Modal In-Context Learning}

In the vision-language domain, a vision encoder paired with a frozen language model demonstrates multi-modal few-shot learning capabilities after training on image-caption datasets, as shown by the Frozen model \citep{tsimpoukelli2021frozen}. Extending this, Flamingo integrates a vision encoder with large language models (LLMs) for enhanced in-context learning across multi-modal tasks, leveraging large-scale web corpora \citep{alayrac2022flamingo}. Similarly, Kosmos-1 exhibits zero-shot, few-shot, and multi-modal chain-of-thought prompting abilities \citep{huang2023kosmos}. METALM introduces a semi-causal language modeling objective to achieve strong ICL performance across vision-language tasks \citep{hao2022language}. The ICL-D3IE approach employs a novel in-context learning framework that iteratively updates diverse demonstrations—including hard, layout-aware, and formatting demonstrations to train large language models (LLMs) for enhanced document information extraction (DIE)\citep{he2023icld3ie}. Recent advancements include creating instruction tuning datasets from existing vision-language tasks or with advanced LLMs like GPT-4, connecting LLMs with powerful vision foundational models like BLIP-2 for multi-modal learning \citep{xu2022multiinstruct,li2023otter,liu2023llava,zhu2023minigpt4,dai2023instructblip}.


\subsection{Speech In-Context Learning}
In the speech area, ~\citet{wang2023neural} treated text-to-speech synthesis as a language modeling task. 
They use audio codec codes as an intermediate representation and propose the first TTS framework with strong in-context learning capability. 
Subsequently, VALLE-X~\citep{zhang2023valle-x} extend the idea to multi-lingual scenarios, demonstrating superior performance in zero-shot cross-lingual text-to-speech synthesis and zero-shot speech-to-speech translation tasks.

\subsection{Author Contributions}
\label{app:authers}
Qingxiu Dong led the project. Qingxiu Dong, Lei Li, Damai Dai, and Ce Zheng discussed and wrote the initial draft of the paper and the second version update. Jingyuan Ma, Rui Li, Heming Xia, and Qingxiu Dong discussed and wrote the third version update, incorporating new work that emerged over the year. Tianyu Liu was responsible for the fourth version update, adding new relevant work and polishing the manuscript. Jingjing Xu, Zhiyong Wu, and Lei Li (CMU) contributed to the revisions and discussions of the first two versions of the paper. Baobao Chang, Xu Sun, Lei Li (CMU), and Zhifang Sui guided the research of Qingxiu Dong, Lei Li, Damai Dai, and Ce Zheng.


%
\subsection{Comparison with other survey papers}
\label{app:comparison_with_paper}
Our survey was drafted and posted on the Arxiv at the end of 2022, which is, to the best of our knowledge,  the very first to review in-context learning in the field. 
We also regularly update this survey in a timely manner, with four major revisions.

Starting from 2023, we notice the emerge of several related survey in the field of in-context learning.
\citet{xu2024context} made a comprehensive review on the choices for models, training procedures and inference algorithms to retrieve demonstrative examples of in-context learning.
\citet{li2023practical} provided practical suggestions on prompt engineering for in-context learning. 
\citet{zhou2023mystery} and \citet{highmore2024context} focused on the theoretical interpretation and analysis of ICL, which corresponds to Section~\ref{sec:analysis} in this survey.
All the above-mentioned survey papers differ with ours in terms of scope and topics. This survey focused on the general development of ICL, including the formal definition of ICL, training strategies, prompt designing strategies, analysis and applications.

\end{document}